\documentclass[10pt, final, conference, letterpaper, onecolumn, oneside]{IEEEtran}

\usepackage{cite}
\usepackage{graphicx}
\usepackage{amsmath}
\usepackage{amssymb}
\usepackage{mathtools}
\usepackage{bm}
\usepackage{epsfig}
\usepackage{times}
\usepackage{color}
\usepackage{url}
\usepackage{hyperref}
\usepackage{array}
\usepackage{multirow}
\usepackage{rotating}
\usepackage{extarrows}
\usepackage{mathabx}
\usepackage{wasysym}
\usepackage{ulem}
\usepackage{braket}
\usepackage{pifont}
\usepackage{soul}
\usepackage{enumitem}
\usepackage{float}
\usepackage{tikz}
\usepackage{titlesec}
\usepackage{booktabs}
\usepackage{subcaption}

\usetikzlibrary{positioning}
\titleclass{\subsubsubsection}{straight}[\subsection]
\newcounter{subsubsubsection}[subsubsection]
\renewcommand{\thesubsubsubsection}{\thesubsubsection.\arabic{subsubsubsection}}
\titleformat{\subsubsubsection}
{\normalfont\normalsize\bfseries}{\thesubsubsubsection}{1em}{}
\titlespacing*{\subsubsubsection}
{0pt}{3.25ex plus 1ex minus .2ex}{1.5ex plus .2ex}

\newcolumntype{P}[1]{>{\centering\arraybackslash}p{#1}}

\begin{document}
\title{Latent Space Characterization of Autoencoder Variants}%
\author{\IEEEauthorblockN{Anika Shrivastava, Renu Rameshan, and Samar Agnihotri}\\%
Correspondence Email: samar.agnihotri@gmail.com%
}

\maketitle
\begin{abstract}
Understanding the latent spaces learned by deep learning models is crucial in exploring how they represent and generate complex data. Autoencoders (AEs) have played a key role in the area of representation learning, with numerous regularization techniques and training principles developed not only to enhance their ability to learn compact and robust representations, but also to reveal how different architectures influence the structure and smoothness of the lower-dimensional non-linear manifold. We strive to characterize the structure of the latent spaces learned by different autoencoders including convolutional autoencoders (CAEs), denoising autoencoders (DAEs), and variational autoencoders (VAEs) and how they change with the perturbations in the input. By characterizing the matrix manifolds corresponding to the latent spaces, we provide an explanation for the well-known observation that the latent spaces of CAE and DAE form non-smooth manifolds, while that of VAE forms a smooth manifold. We also map the points of the matrix manifold to a Hilbert space using distance preserving transforms and provide an alternate view in terms of the subspaces generated in the Hilbert space as a function of the distortion in the input. The results show that the latent manifolds of CAE and DAE are stratified with each stratum being a smooth product manifold, while the manifold of VAE is a smooth product manifold of two symmetric positive definite matrices and a symmetric positive semi-definite matrix.
\end{abstract}

\section{Introduction}
\label{sec:introduction}
With the emergence of cutting-edge deep learning models, the field of image processing has seen significant progress. However, this advancement necessitates a deeper understanding of the inner workings of these models, specifically how they represent data. Autoencoders, introduced in \cite{rumelhart1986learning}, serve as the foundation for a wide range of unsupervised learning models \cite{zhai2018autoencoder} and have gained significant attention for their ability to learn meaningful representations of data. They learn these representations with the help of a simple end-to-end structure involving two main components: an encoder and a decoder. The input \(y \in \mathbb{R}^D \) is mapped to a latent representation \(z \in \mathbb{R}^d \) via an encoding function \( f: \mathbb{R}^D \rightarrow \mathbb{R}^d \), and then the decoder reconstructs it back in the original space using a decoding function \( g: \mathbb{R}^d \rightarrow \mathbb{R}^D \), minimizing the reconstruction loss \( \mathcal{L}(\mathbf{y}, \mathbf{\hat{y}}) \), where \(\mathbf{y}\) is the original input and \(\mathbf{\hat{y}}\) is its reconstruction. In essence, the latent space is where \(z\) lies. Characterizing the latent space involves analyzing how autoencoders arrange data within this space, understanding the properties of this space, and assessing whether smooth navigation is possible within the space. We believe that knowing the structure of the latent space can guide one in designing better restoration algorithms. 

Traditionally, autoencoders were introduced as a dimensionality reduction technique, where the latent space had a dimension \( d < D \), resulting in an under-complete autoencoder. This dimensionality restriction acted as a form of regularization, forcing the model to learn only the most important features of \(y\). However, some variants of autoencoders, known as over-complete autoencoders, employ latent spaces with dimensions equal to or even larger than the input space. While this design has the potential to capture the closest reconstruction of the input image, it also introduces the risk of the model learning an identity function \cite{bengio2013representation}, where it simply replicates the input, thus failing to learn any useful representations. To prevent this, over-complete models are often combined with regularization techniques such as weight decay, adding noise to input images \cite{vincent2008extracting}, imposing sparsity constraints {\cite{ng2011sparse}}, or by adding a penalty term to the loss function to make the space contractive {\cite{rifai2011contractive}}. These regularizations help in structuring the latent space to be compact and robust against small variations in the input data, enabling the model to learn robust and meaningful patterns rather than merely copying the input. Additionally, some variants introduce a stochastic component by enforcing a probabilistic latent space, which ensures smooth latent manifold leading to better generalization \cite{kingma2013auto}. In Section \ref{sec:lit.survey}, we discuss how these regularization methods shape the properties of the latent space. However, while these methods impose some structure on the latent space, they do not directly explain the underlying manifold---specifically, its geometry and properties. Our work aims to bridge this gap by providing a more detailed understanding of the manifold structure learned by different autoencoder variants.

We aim to characterize the latent spaces of overcomplete \textbf{Convolutional autoencoders (CAE)}, \textbf{Denoising autoecnoders (DAE)}, and \textbf{Variational autoencoders (VAE)} by analyzing how varying levels of noise impact their respective learned latent manifolds and whether the structures of these spaces permit smooth movement within them. Empirically, it is observed that autoencoders exhibit a non-smooth latent structure \cite{oring2021autoencoder}, while VAEs tend to have a smooth latent structure \cite{cristovao2020generating}. A simple experiment to visually illustrate this difference involves interpolating between two data points by decoding convex combinations of their latent vectors \cite{berthelot2018understanding}. For CAE and DAE, this often leads to artifacts or unrelated output, indicating the lack of smooth transitions between the two points. In contrast, VAE exhibits a coherent and smooth transition, reflecting its continuous latent space. Our approach builds upon the	work of \cite{sharma2021distance}, where video tensors are modeled as points on the product manifold (PM) formed by the Cartesian product of symmetric positive semi-definite (SPSD) matrix manifolds. We adapt this method for the encoded tensors extracted from each model's latent space and examine the ranks of the SPSD matrices to analyze the structure of the learned latent manifold. This analysis provides evidence for the fact that the latent spaces of CAE and DAE have non-smooth structure as those are stratified manifolds with each stratum being a smooth manifold based on the ranks, while that of the VAE forms a smooth product manifold of SPD and SPSD matrices. Furthermore, we transform these PM points to the Hilbert space using a distance based positive-definite kernel \cite{sharma2021distance}, allowing us to analyze the latent spaces in terms of subspaces. 

Our main contribution is in characterizing the manifold by using a simple observation namely, the latent tensors lie on a product manifold of symmetric positive semidefinite matrices. We also explore how the manifold structure changes with perturbations in the input. Perturbations are modeled by additive white Gaussian noise with different variances. We show that while CAE and DAE have a stratified matrix manifold, VAE has a matrix manifold that is smooth. 

\noindent\textbf{Organization:} The remainder of the paper is structured as follows. Section~\ref{sec:lit.survey} provides a brief literature review. Section~\ref{sec:approach} discusses the approach used for the characterization of latent spaces, followed by experimental details in Section~\ref{sec:details}. Section~\ref{sec:results} analyzes the results obtained. Finally, Section~\ref{sec:conc} concludes the paper.

\section{Related Work}
\label{sec:lit.survey}
\textbf{Regularization-guided latent spaces.}
The widely recognized manifold hypothesis \cite{fefferman2016testing} suggests that a finite set of high dimensional data points concentrate near or on a lower-dimensional manifold \(\mathcal{M}\). A manifold is basically a topological space that locally resembles Euclidean space near each point, and autoencoders are instrumental in learning this underlying latent manifold. Several autoencoder variants employ regularization techniques to enhance the robustness and structure of the underlying latent manifold.  
\cite{vincent2008extracting} introduced Denoising autoencoders (DAEs), a modification to the traditional autoencoders where the model learns to reconstruct clean images \(\hat{y}\) from noisy/corrupted inputs \(\tilde{y}\), thereby, minimizing the reconstruction loss \( \mathcal{L}(y, \hat{y}) \). From a manifold learning perspective, the latent space of DAEs identifies the lower dimensional manifold where the clean data resides and DAEs learn to map the corrupted data back onto this manifold. This enables the model to generalize better, capturing essential latent representations while being robust to noise. 
Based on a similar motive of learning the lower-dimensional manifold and robust latent representations \cite{rifai2011contractive} add a contractive penalty to the learning process. Unlike traditional autoencoders, contractive autoencoders apply a regularization term to the encoder’s Jacobian matrix, penalizing the sensitivity of the latent space to small input changes. In other words, the underlying latent manifold becomes locally invariant to small variations in the input and contracts the latent space along these directions of unimportant variations. Similarly, sparse autoencoders \cite{ng2011sparse} learn non-redundant representations by enforcing sparsity in the hidden units. By activating only a few neurons at a time, the model captures more distinct, disentangled features, resulting in a sparse, interpretable and efficient latent space. 
In addition to these techniques, Variational autoencoders (VAEs) \cite{doersch2016tutorial} introduce a probabilistic structure to the latent space by learning a distribution over the latent variables, rather than representing them as fixed points. This pushes the latent space toward being continuous and smooth, facilitating tasks like data generation and interpolation. 

\noindent\textbf{Representation geometry.} Several studies explore and regularize the geometry of the latent representation in VAEs. For instance, \cite{chadebec2022geometric} show that the latent manifold learned by VAEs can be modeled as a Riemannian manifold, while \cite{chen2020learning} extend the VAE framework to learn flat latent manifolds by regularizing the metric tensor to be a scaled identity matrix. \cite{connor2021variational} incorporate a learnable manifold model into the latent space to bring the prior distribution closer to the true data manifold. Additionally,
{\cite{leeb2022exploring}} develop tools to exploit the locally contractive behaviour of VAEs to better understand the learned manifold. These and many other studies assume that VAEs learn a smooth manifold, whereas AEs learn a non-smooth manifold \cite{oring2021autoencoder}, but the exact structure and properties of these manifolds have not been thoroughly explored.

We aim to precisely capture the structure of the latent space and how it evolves when processing images with varying levels of noise. Our results confirm that the latent manifolds learned by AEs are non-smooth, while the manifold learned by VAEs is smooth - explaining the reasons behind this behavior and characterizing the space in detail. Many studies have demonstrated the effectiveness of modeling sample data as points in the product manifold across various vision tasks \cite{abdelkader2011silhouette,lui2010action,lui2012human,sharma2019linearized}. Motivated by this, we strive to thoroughly model the latent space points in the PM of the SPSD matrices to characterize the behaviour of latent spaces of different models.

\section{Product manifold structure}
\label{sec:approach}
In this section, we describe the details of the autoencoder network used for feature extraction and the method we adopt for modeling the encoded latent tensors as points in the PM of SPSD matrices, and for further transforming PM points to the Hilbert space.

\subsection{Model Architectures} 
\begin{figure*}[!t]
    \centering
    \includegraphics[width=\textwidth]{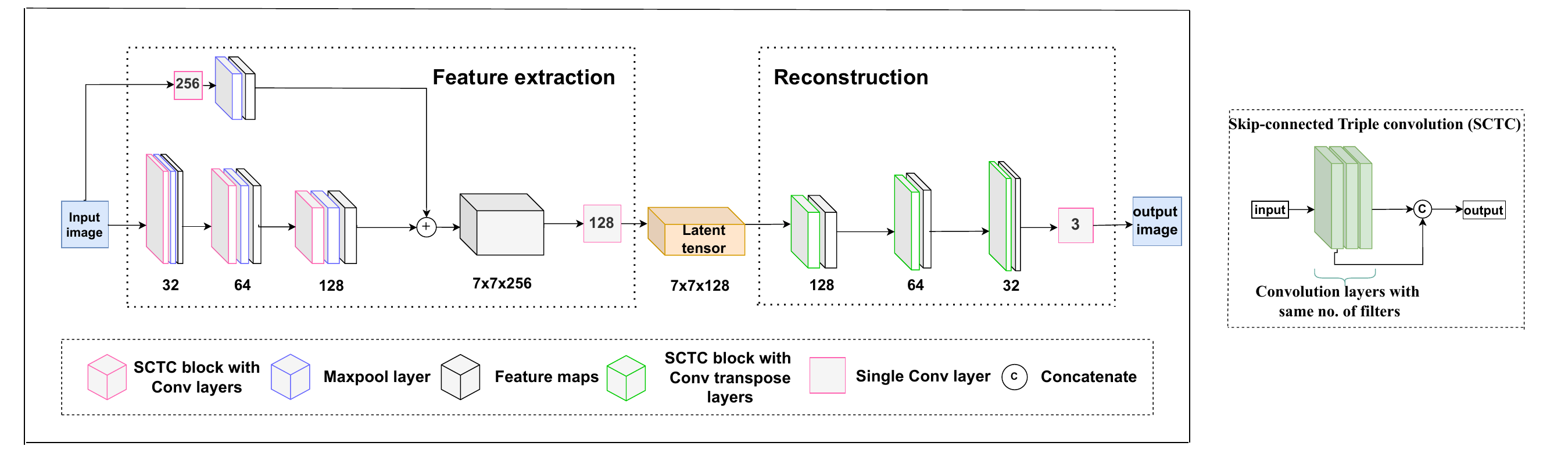}
    \caption{The SCTC block used in CAE and DAE models.}
    \label{fig:ae}
\end{figure*}

The architecture used for extracting latent tensors in both CAE and DAE models is built of "Skip-Connected Triple Convolution” (SCTC) block as shown in Fig. \ref{fig:ae}. Each SCTC block contains three convolutional layers with the same number of filters and a skip connection from the first convolution to the third convolution. The encoder is composed of three such layers, each followed by max-pooling. Additionally, a skip connection is introduced directly from the input image to the latent representation using a single convolutional and max-pooling layer. The decoder mirrors the encoder's structure, using the SCTC blocks with transpose convolution layers to reconstruct the images from the latent tensor. We select the SCTC blocks after extensive experimentation. To assess the impact of the SCTC blocks, we replace them with standard convolution layers, which result in reduced PSNR, confirming their importance in preserving image details. Additionally, removing the direct skip connection from the input to the latent tensor again leads to a drop in PSNR, underscoring their role in better feature retention.

For VAEs, we use a standard convolutional VAE architecture. However, instead of using linear layers to compute the mean \((\mu)\) and log variance, we employ convolution layers to generate a latent tensor instead of a latent vector. Through experimentation, we confirm that both latent vectors and latent tensors yield similar reconstruction output. Based on this, we opt for latent tensors to maintain consistency, with the shape of the extracted latent tensor fixed at \(7\times7\times128\) for all models to ensure fair comparison.

\subsection{Latent tensors as points in a product manifold}
\label{sec:PM}
\begin{figure*}[!t]
    \centering
    \includegraphics[width=\textwidth]{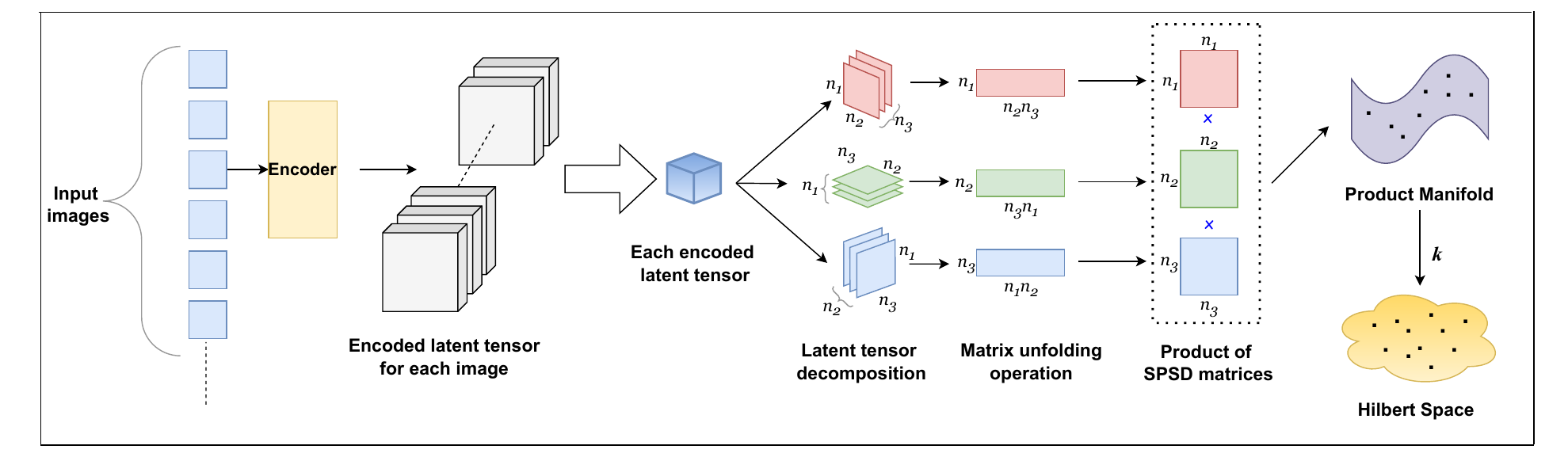}
    \caption{Pipeline of the proposed approach.}
    \label{fig:pipeline}
\end{figure*}

The encoded latent tensors can be understood as points lying on a PM of the SPSD matrices. An illustration of the pipeline used for representing encoded latent tensors as points in the PM of the SPSD matrices is shown in Fig. \ref{fig:pipeline} (inspired by \cite{sharma2021distance}). Let the encoded feature tensors have the shape \((N, n_1, n_2, n_3)\), where \(N\) represents the number of test samples. Each feature tensor can be interpreted as a point \( F \in \mathbb{R}^{n_1 \times n_2 \times n_3} \), with \(n_1\), \(n_2\) and \(n_3\) corresponding to height, width, and number of channels of the encoded image, respectively. These tensors are then decomposed into a set of three matrices \( F \mapsto \{F^{(1)}, F^{(2)}, F^{(3)}\} \), using matrix unfolding, where \( F^{(1)} \in \mathbb{R}^{n_1 \times (n_2 \cdot n_3)} \), \( F^{(2)} \in \mathbb{R}^{n_2 \times (n_3 \cdot n_1)} \), and \( F^{(3)} \in \mathbb{R}^{n_3 \times (n_1 \cdot n_2)} \). For each \(F^{(i)}\), a covariance matrix is calculated denoted as \(S^{(1)}, S^{(2)}, S^{(3)}\) and these are inherently the SPSD matrices. The Cartesian product of these covariance matrices is a product manifold of the SPSD manifolds \cite{rodola2019functional}.

By definition, the SPSD manifold \(S_+^n(r)\) \cite{bonnabel2010riemannian} is the space of \(n \times n\) SPSD matrices of fixed rank \(r\). The SPSD matrices sharing the same rank belong to the same manifold. The collection of all \(n\times n\) SPSD matrices with rank \(\le r\) is not a manifold.  It is well known that the collection of all $n \times n$ SPSD matrices with varying ranks, forms a stratified manifold \cite{massart2019curvature}. The ranks \(r_1, r_2, r_3\) of the matrices \(S^{(1)}, S^{(2)}, S^{(3)}\), respectively, form a tuple \((r_1, r_2, r_3)\), characterizing the overall rank configuration of the latent tensor within the PM. We show in Section \ref{sec:results} that the way this rank tuple behaves with varying noise levels is different for the three architectures. The variability in these ranks indicate whether the underlying manifold is smooth or stratified. 

\subsection{Transformation to Hilbert space}
\label{sec:HS}
To simplify the understanding, instead of viewing the latent representation as a tensor in the SPSD manifold, we adopt an alternative approach by embedding these points into a Hilbert space. Each covariance descriptor \(S^{(i)}\) is regularized to a fixed rank \(r_i\) by replacing zero eigenvalues with small epsilon value, where \(r_i\) corresponds to the maximum rank observed across all test samples for each \( i \in \{1, 2, 3\} \).
The decomposition of each \(S^{(i)}\) is given as \cite{bonnabel2010riemannian}:
\begin{equation} 
S^{(i)} = A^{(i)}A^{(i)\top} = (U^{(i)}R^{(i)})(U^{(i)}R^{(i)})^{\top} = U^{(i)}R^{(i)2}U^{(i)\top},
\end{equation}
for \( i \in \{1, 2, 3\} \) corresponding to each unfolding. Here, \(U \in \mathbb{R}^{n\times r}\) has orthonormal columns; \(n\) is the size of \(S\) and \(r\) its rank. \(R\) is an SPD matrix of size \(r\).  Following \cite{sharma2021distance}, the geodesic distance function between any two points \(\gamma_1, \gamma_2\) on the PM of the SPSD matrices is defined as:

\begin{equation}
d^2_g(\gamma_1, \gamma_2) = \sum_{i=1}^{3} \Bigg( \frac{1}{2} \|U_1^{(i)}U_1^{(i)T} - U_2^{(i)}U_2^{(i)T}\|_F^2 + \lambda^{(i)} \|\log(R_1^{(i)}) - \log(R_2^{(i)})\|_F^2 \Bigg).
\label{eq:dist}
\end{equation}

For further analysis we use the positive definite linear kernel function that follows from Eq. \ref{eq:dist}:\begin{equation}
k_{lin}(\gamma_1, \gamma_2) = \sum_{i=1}^{3} w_i \Bigg( \|U_1^{(i)T}U_2^{(i)}\|_F^2 + \lambda^{(i)} \text{tr}\left( \log(R_1^{(i)}) \log(R_2^{(i)}) \right) \Bigg),
\label{eq:kernel}
\end{equation}
where \(w_{i}\) denotes weight for each factor manifold and \(tr\) denotes trace of a matrix.
The transformation from the PM to the Hilbert space is achieved using this distance based positive definite kernel function. It has been  shown in  \cite{sharma2021distance} that such a kernel ensures that the distances between points in the manifold are maintained in the Hilbert space after the transformation. 

Using the kernel in Eq. \ref{eq:kernel}, we can obtain virtual features (VF) for each of the data tensor as described in  \cite{zhang2012scaling}. If there are $N$ data points, then the virtual feature is a length $N$ vector obtained from the kernel gram matrix ($K$) and its diagonalization. Following the observation that not all the eigenvalues of $K$ are significant, we do a dimensionality reduction and map the manifold points to a lower dimensional subspace of $\mathbb{R}^N$. In Section \ref{sec:results}, we demonstrate how the dimensionality of the space changes with varying noise levels for the three autoencoder variants.

\section{Experimental Setup}
\label{sec:details}
\textbf{Training data:} We train our models on the MNIST dataset. The CAE and VAE models are trained using 60,000 clean images, while the DAE is trained on a noisy version of the dataset, with Gaussian noise (sigma = 0.05) added to the training images.

\noindent\textbf{Testing data:} To effectively capture the changes in structure of the underlying latent manifold for each model, we construct a comprehensive test dataset from the MNIST test set. This dataset includes multiple classes, each containing 300 images. The first class contains clean images, while the subsequent classes are progressively corrupted with Gaussian white noise, with the variance increasing in increments of 0.01.

\noindent\textbf{Training loss:} For the CAE and DAE, we employ a custom loss function that combines the weighted sum of MSE and SSIM losses, with equal weights. For the VAE, we train it using the MSE reconstruction loss along with the Kullback–Leibler Divergence loss, where the KLD term is weighted by a parameter \(\beta=0.8\).

\section{Results and analysis}
\label{sec:results}
Empirical observations in the existing literature show that autoencoders like CAEs and DAEs tend to exhibit a non-smooth latent structure, while VAEs are known for producing a smooth latent structure.
We aim to explain this widely discussed hypothesis by exploring what these manifolds exactly are  and motivate our findings from different perspectives.

\subsection{Latent space structure in manifold space}
\begin{figure*}[!t]
    \centering
    \includegraphics[width=0.48 \textwidth]{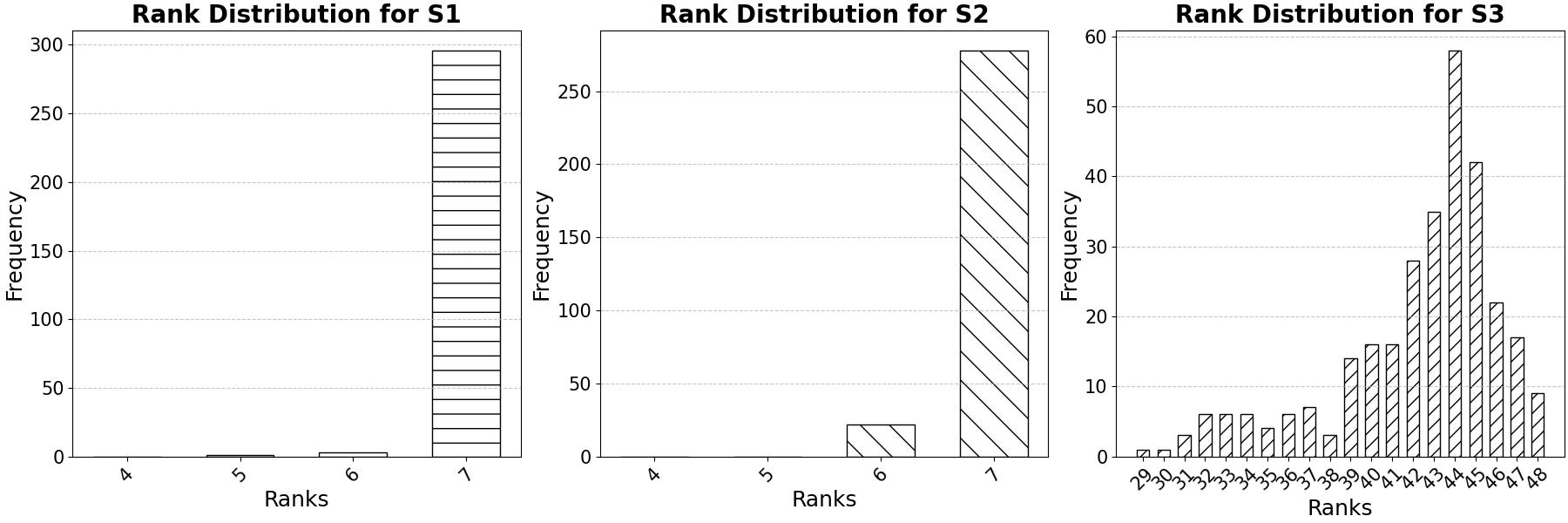} \quad \includegraphics[width=0.48 \textwidth]{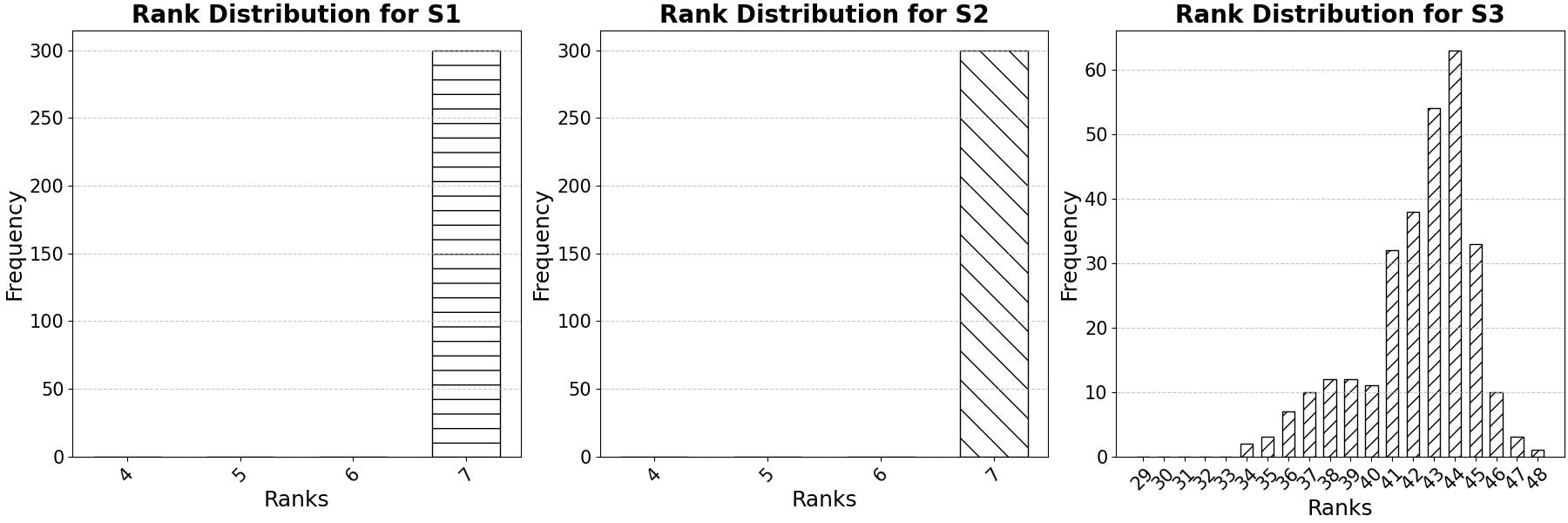} \\
\includegraphics[width=0.48 \textwidth]{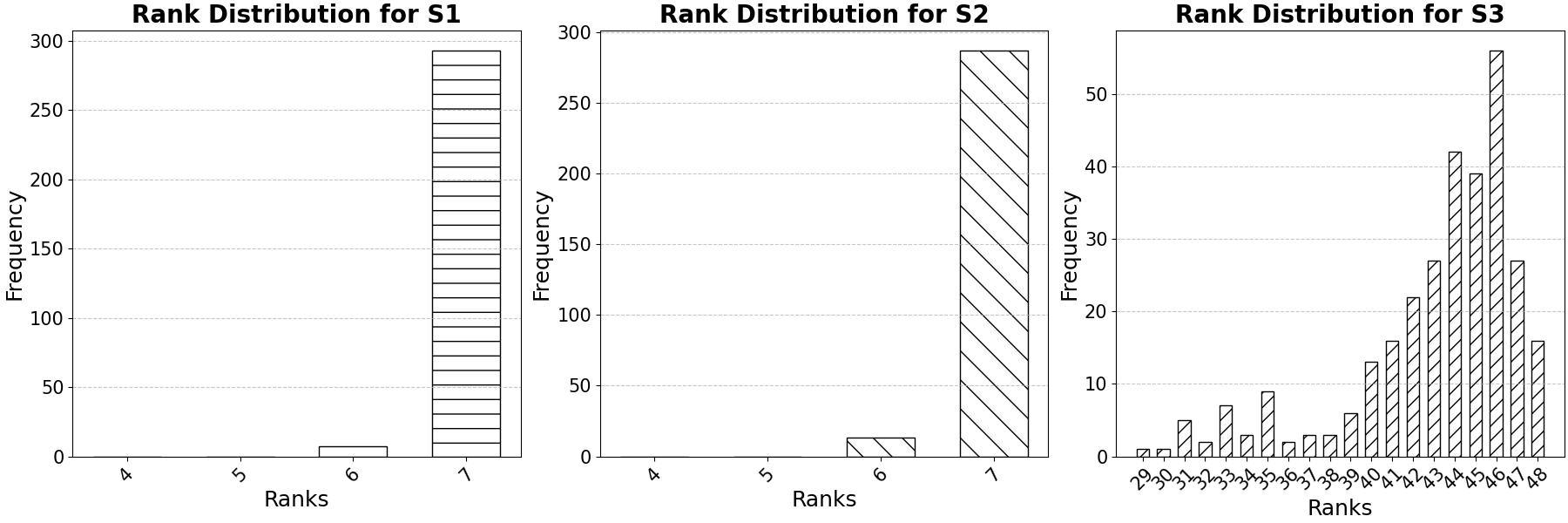} \quad \includegraphics[width=0.48 \textwidth]{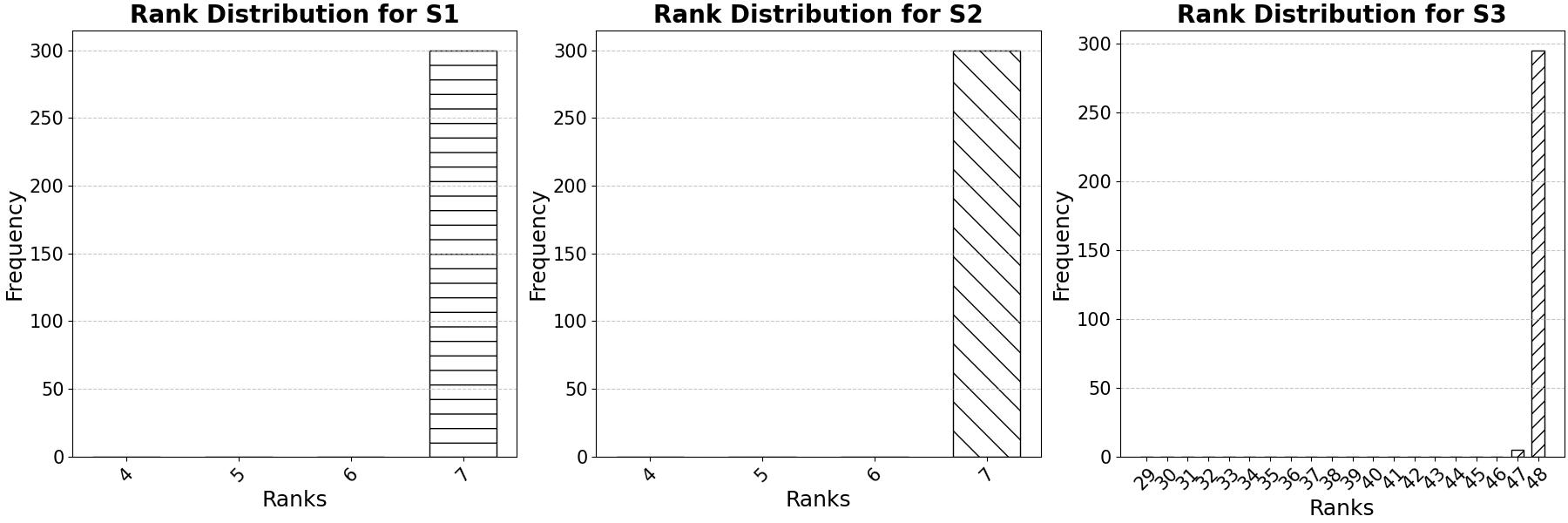} \\ 
\includegraphics[width=0.48 \textwidth]{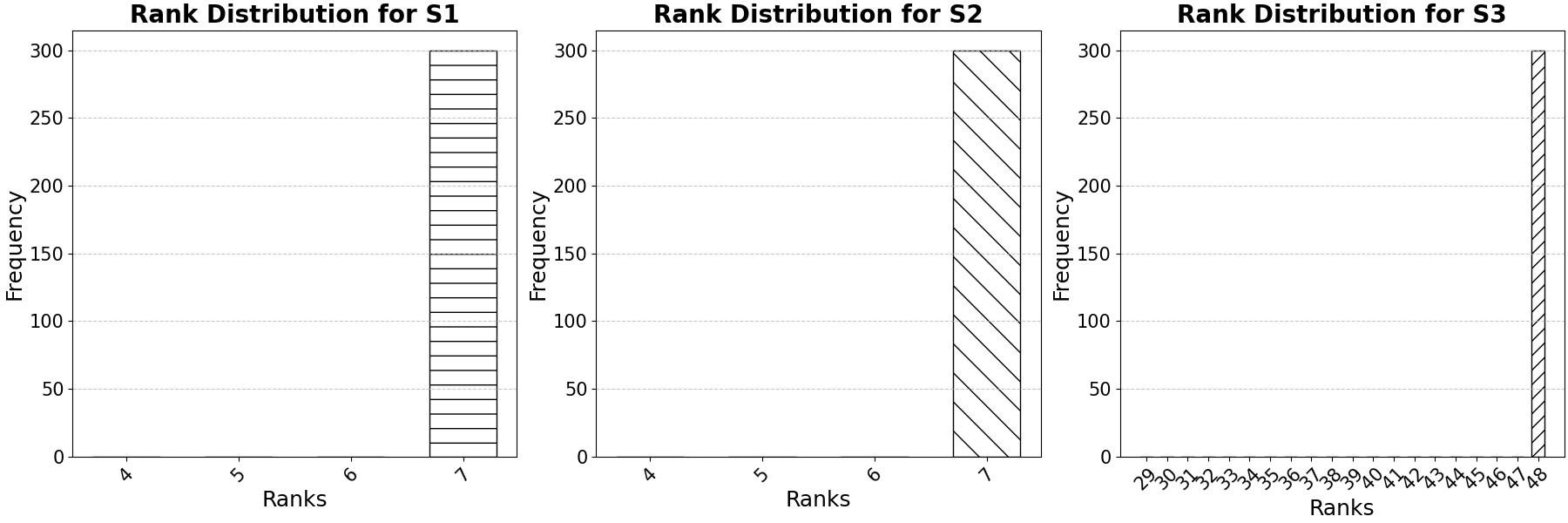} \quad \includegraphics[width=0.48 \textwidth]{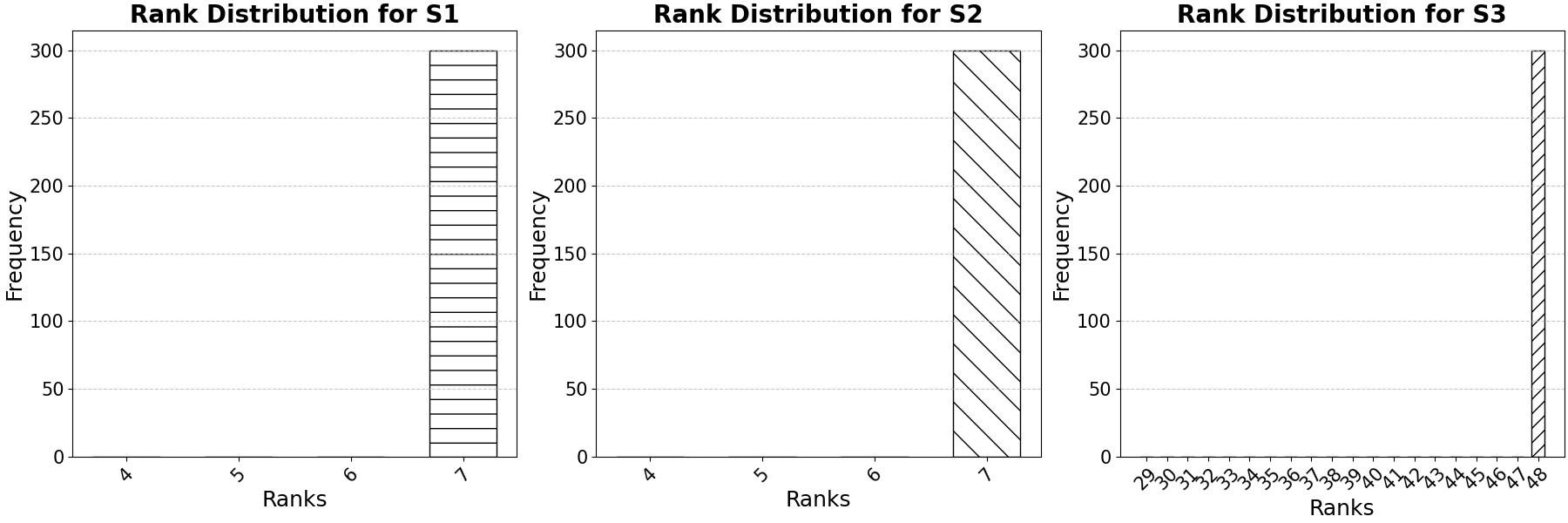} \\

    \caption{Histograms of ranks of $S^{(1)},\,S^{(2)},\,S^{(3)}$ for the three models on $300$ test samples. Left side is for clean and right for noisy with standard deviation $0.1$. From top to bottom: CAE, DAE, VAE.}
    \label{fig:hist}
\end{figure*}

In our first experiment, we use the Berkeley segmentation dataset (BSDS) to train the CAE and DAE models. The latent tensor corresponding to each image in the test dataset is modeled as a point on the PM formed of the SPSD matrices, as described in subsection~\ref{sec:PM}. \(S^{(1)}, S^{(2)}, S^{(3)}\) have shapes \(32\times 32\),  \(32\times 32\) and \(128\times 128\), respectively, corresponding to each unrolled matrix. The ranks of these SPSD matrices are then calculated across different noise levels. The rank configuration ($r_1, r_2, r_3$) decides the product manifold in which the latent tensor lies. It is observed that all the latent tensors do not lie on the same product manifold in the case of CAE and DAE. For clean images and at lower noise levels, the latent tensors are distributed across different strata of the product manifold whereas at higher noise levels they tend to lie on fewer strata. For a fair comparison between the three models, we switch to the MNIST data as all three models present a similar reconstruction performance on this dataset. The CAE and DAE exhibit a behaviour similar to that for the BSDS. Contrary to this behaviour, the latent tensors of the VAE lie in the same product manifold for clean as well as for noisy cases at all noise variances. The rank variablilty across the three models is given in Fig. \ref{fig:hist}. The ranks of the three SPSD matrices for varying noise are presented in Table \ref{tab2}. These ranks are reported as ranges (min, max) for each case to reflect the observed variability across the dataset. We observe that the ranks of the SPSD matrices, mostly $S^{(3)}$, for CAE and DAE vary across different noise levels, while for VAE, the corresponding ranks remain fixed.

The variability in the ranks of the SPSD matrices observed in CAE and DAE results in a \textbf{stratified manifold}. Since each SPSD matrix lies on a specific SPSD manifold defined by its rank, the latent spaces of the CAE and DAE span multiple product manifolds determined by the tuple \((r_1, r_2, r_3)\) creating stratification of space \cite{takatsu2011wasserstein}. In contrast, the VAE shows consistent ranks across all noise levels for the three covariance matrices, indicating that all tensors lie on the same product manifold - product of two SPD manifolds and an SPSD manifold. This consistency results in smooth movement within the latent space from one point to another, describing the latent space of the VAE as a \textbf{smooth product manifold}. 

\subsection{Latent space structure in Hilbert space}
\begin{figure*}[!t]
    \centering
    \includegraphics[width=\textwidth]{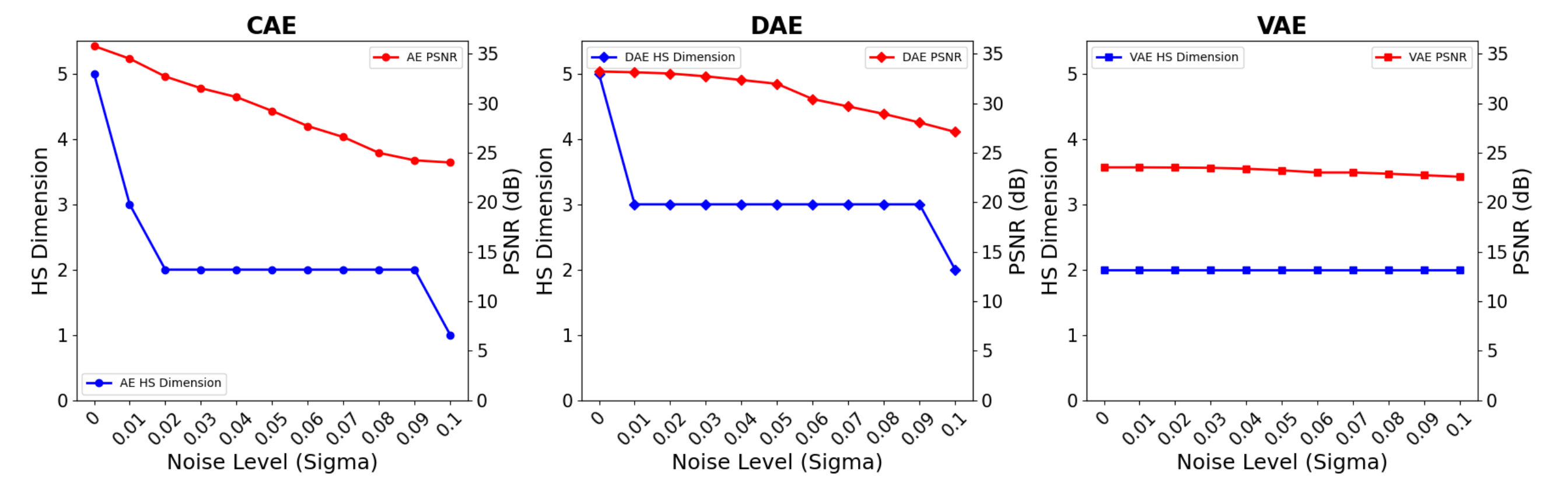}
    \caption{The Hilbert space dimensionality and PSNR versus noise level for CAE, DAE, and VAE.}
    \label{fig:dim_psnr}
\end{figure*}

The stratified structure observed in CAE and DAE can be difficult to visualize directly within the product manifold space. Therefore, we transform the PM points into Hilbert space, as detailed in subsection~\ref{sec:HS}. This is justified as the distance between the points are preserved in both the spaces as shown in \cite{sharma2021distance}. For a test dataset with $N$ data points, an  \(N \times N\) kernel-gram matrix is generated. We proceed by calculating virtual features, followed by dimensionality reduction to derive N-length vectors lying on a \(d\)-dimensional subspace of $\mathbb{R}^N$. The dimension \(d\) for each noise level is determined by minimizing the difference between the original kernel-gram matrix and its rank-\(i\) approximations and selecting the smallest \(i\) such that the norm falls below a certain threshold. Fig. \ref{fig:dim_psnr} illustrates the change in \(d\) with noise levels.

\begin{table*}
\caption{Ranks of unrolled covariance matrices across different noise level for CAE, DAE, and VAE.}
\centering
\label{tab2}
\resizebox{\textwidth}{!}{
\begin{tabular}{>{\centering}m{0.9cm} >{\centering}m{4.6cm} >{\centering}m{4.6cm} >{\centering\arraybackslash}m{4.6cm}}
	\toprule
	\textbf{Noise levels} & \textbf{CAE (latent shape: 7x7x128)} & \textbf{DAE (latent shape: 7x7x128)} & \textbf{VAE (latent shape: 7x7x128)} \\
	\midrule
zero & S1: (5, 7), S2: (6, 7), S3: (29, 48) & S1: (6, 7), S2: (6, 7), S3: (29, 48) & S1: (7, 7), S2: (7, 7), S3: (48, 48) \\
0.01 & S1: (6, 7), S2: (7, 7), S3: (30, 48) & S1: (7, 7), S2: (6, 7), S3: (42, 48) & S1: (7, 7), S2: (7, 7), S3: (48, 48) \\
0.02 & S1: (5, 7), S2: (7, 7), S3: (30, 48) & S1: (6, 7), S2: (7, 7), S3: (42, 48) & S1: (7, 7), S2: (7, 7), S3: (48, 48) \\
0.03 & S1: (6, 7), S2: (7, 7), S3: (32, 48) & S1: (6, 7), S2: (7, 7), S3: (42, 48) & S1: (7, 7), S2: (7, 7), S3: (48, 48) \\
0.04 & S1: (6, 7), S2: (7, 7), S3: (32, 48) & S1: (7, 7), S2: (7, 7), S3: (43, 48) & S1: (7, 7), S2: (7, 7), S3: (48, 48) \\
0.05 & S1: (6, 7), S2: (7, 7), S3: (31, 48) & S1: (7, 7), S2: (7, 7), S3: (44, 48) & S1: (7, 7), S2: (7, 7), S3: (48, 48) \\
0.06 & S1: (7, 7), S2: (7, 7), S3: (31, 48) & S1: (7, 7), S2: (7, 7), S3: (43, 48) & S1: (7, 7), S2: (7, 7), S3: (48, 48) \\
0.07 & S1: (7, 7), S2: (7, 7), S3: (32, 48) & S1: (7, 7), S2: (7, 7), S3: (45, 48) & S1: (7, 7), S2: (7, 7), S3: (48, 48) \\
0.08 & S1: (7, 7), S2: (7, 7), S3: (33, 48) & S1: (7, 7), S2: (7, 7), S3: (46, 48) & S1: (7, 7), S2: (7, 7), S3: (48, 48) \\
0.09 & S1: (7, 7), S2: (7, 7), S3: (33, 48) & S1: (7, 7), S2: (7, 7), S3: (47, 48) & S1: (7, 7), S2: (7, 7), S3: (48, 48) \\
0.1  & S1: (7, 7), S2: (7, 7), S3: (34, 48) & S1: (7, 7), S2: (7, 7), S3: (47, 48) & S1: (7, 7), S2: (7, 7), S3: (48, 48) \\
\bottomrule
\end{tabular}
}
\end{table*}

Rather than dealing with multiple product manifolds that arise for CAE and DAE, we use a regularizer - adding a small value to all the zero eigenvalues of the SPSD matrices - and push all the points to lie on a single product manifold. While it may seem that this simplification destroys the structure, our results show that the rank variability gets reflected as variability of subspace dimension in the Hilbert space.   

We observe that for the CAE and DAE, the dimensionality of subspaces decreases as the input transitions from clean to noisy, indicating that the subspaces in the Hilbert space change with increasing noise. The CAE experiences a sharper drop in dimensionality, while the DAE preserves it slightly better. In contrast, the VAE points lie in the same subspace regardless of the noise level.

\begin{figure*}[!t]
    \centering
    \includegraphics[width=\textwidth]{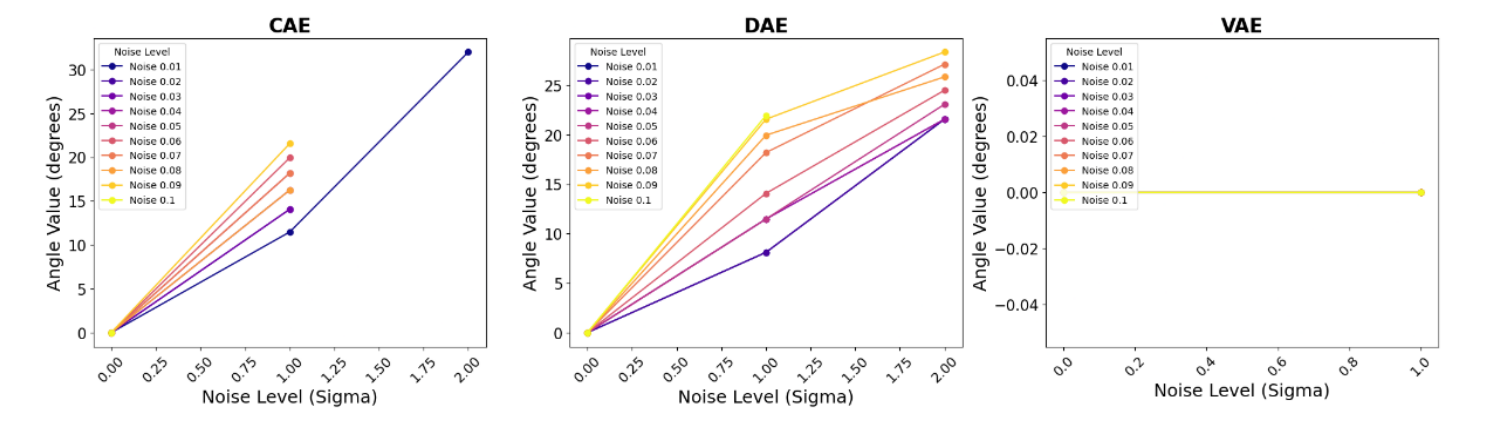}
    \caption{Principal angle variations for CAE, DAE, VAE}
    \label{fig:angles}
\end{figure*}

With noisy subspace dimensions differing from those of the clean subspace, we have so far established that CAE and DAE points lie on  distinct subspaces for noisy cases.  To examine how the subspaces corresponding to noisy inputs are oriented with respect to the clean ones, we calculate the \textit{principal angles} \cite{knyazev2012principal} between noisy and clean subspaces at each noise level. Given two subspaces \( \mathcal{X} \) and \( \mathcal{X'} \) with dimensions \( d \) and \( d' \), respectively, the number of principal angles is determined by \( m = \min(d, d') \). The results, presented in Fig.\ref{fig:angles}, show that for the CAE and DAE, the  principal angles increase with noise level, suggesting that the noisy subspaces diverge away from the clean ones with noise. This divergence is more pronounced in the CAE. In contrast, the VAE shows zero principal angles as expected. 

We also examine how PSNR behaves across different noise levels (Fig.~\ref{fig:dim_psnr}). It is observed that  as the subspace dimension decreases, the PSNR tends to drop, particularly in CAE and DAE, whereas VAE maintains both constant dimensionality and consistent PSNR across all noise levels, suggesting a connection between the two. 

\subsection{Visualizing the latent tensors}
\begin{figure*}[!t]
    \centering
    \includegraphics[width=\textwidth]{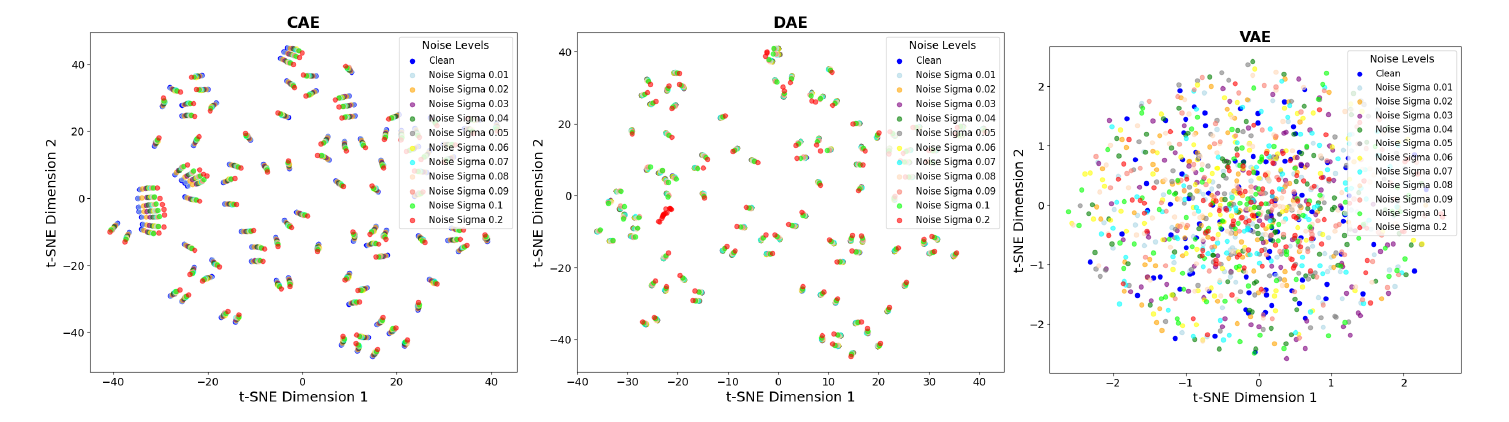}
    \caption{t-SNE plot of latent tensors for CAE, DAE, VAE}
    \label{fig:tsne}
\end{figure*}

To provide a clear visual representation of the learned latent tensors for the three models across different noise levels, we used t-distributed Stochastic Neighbor Embedding (t-SNE) \cite{van2008visualizing} of the latent tensors extracted from each model. The t-SNE is a dimensionality reduction technique that projects high-dimensional data into a two-dimensional Euclidean space, enabling easier visual comparison of latent spaces and complementing the more abstract analysis in the Hilbert space.
Fig. \ref{fig:tsne} shows the t-SNE plots of the flattened latent tensors for each model across different noise levels, with each color representing a different noise level and blue dots denoting the clean images.

We observe that as noise increases, the latent tensors for CAE and DAE diverge further from the clean points, thus confirming our earlier observation from the Hilbert space analysis, where the principal angles between noisy and clean subspaces consistently increased with higher noise levels. In DAE, the divergence is more gradual compared to CAE. In contrast, for VAE,  the t-SNE coordinate range shows  that the divergence between clean and noisy points is quite small, with both clean and noisy representations remaining within a single, tightly clustered region centered around the origin. These observations reinforce the behaviour that we observe in both the manifold and Hilbert space analyses.

\subsection{Discussion}
In this subsection, we discuss the implications of the observations from the results discussed earlier in this section, focusing on their relevance to the denoising application from the latent space perspective.

It is observed that the CAE and DAE have a latent space which is a stratified manifold. While moving within a stratum is smooth, transitioning across strata is not so due to the dimensionality differences among the strata. It is interesting to note that with increasing noise, the manifold becomes a smooth SPD manifold and moving from this higher rank manifold of SPD matrices to the lower rank manifold of SPSD matrices is easier as it only involves thresholding of the smaller eigenvalues. This observation points to a possibility of using both the autoencoders for denoising.

For VAEs, as outlined in \cite[Example 1.34]{lee2012smooth}, the Cartesian product of manifolds is a smooth product manifold if each component manifold is smooth. Since both SPD and SPSD (of fixed rank) are smooth manifolds, the latent space of the VAE can be described as a smooth product manifold. This smooth manifold structure may lead to far simpler denoising algorithms than those with the CAEs and DAEs. Our primary contribution lies in characterizing the latent manifold as a product manifold of the SPD/SPSD matrices, where the core distinction between smooth and non-smooth manifolds is rooted in the structure of this product manifold. The CAE and DAE form stratified product manifolds due to rank variability of their respective SPSD matrices, leading to discontinuities and non-smooth transitions across strata. On the other hand, the VAE, with its uniform rank structure, results in a smooth product manifold.

The Hilbert space analysis for the three autoencoders reveals the following structure. For the CAE and DAE, the subspace representing noisy data rotates away from that of clean data and has a lower dimension. Whereas, for the VAE, the noisy and clean data points are transformed to latent space points that can be mapped to the same Hilbert space. The t-SNE plots also reflect these distinct structural differences in latent spaces of the three autoencoders. The t-SNE plot of VAE feature-points appear as a coherent ball despite varying noise levels. Contrary to this, these points are seen as elongated points in the t-SNE plot for the CAE. The DAE, being midway between these two, has more tightly clustered points.

All the three analysis methods provide strong and concurring evidence for the smoothness of the latent space of VAE and the non-smooth structure of latent space of CAE and DAE.

\section{\uppercase{Conclusion and Future Work}}
\label{sec:conc}
We characterize the latent spaces of different autoencoder models, specifically CAEs, DAEs and VAEs, to gain an understanding of their respective smoothness properties. For this, we explore the latent space in two domains: the manifold space and the Hilbert space. In the manifold space, with the help of a simple observation that the latent tensors lie on a product manifold of the SPSD matrices, we observe the variability in ranks of the SPSD matrices in the CAE and DAE that result in a stratified structure, where each stratum is smooth but the overall structure is non-smooth due to discontinuities among strata. In contrast, the VAE shows consistent ranks, forming a smooth product manifold. In the Hilbert space, varying dimensions and increasing principal angles between clean and noisy subspaces in the CAE and DAE suggest distinct subspaces for clean and noisy data, while the VAE maintains the subspace with the same dimensionality for both, with zero principal angles. We also note a close relationship between subspace dimensionality and reconstruction performance across models. These results are corroborated by the t-SNE plots where significant divergence is observed between clean and noisy points in the CAE and DAE, while such points for the VAE are tightly clustered near the origin.

Much work remains to be done. We plan to extend this analysis to other autoencoder variants and validate the results with more datasets. Further, we also plan to characterize the latent spaces of other generative models. As a constructive next step, we intend to devise denoising and deblurring algorithms by leveraging the understanding of the manifold structure of autoencoders, particularly VAEs.

\end{document}